\documentclass[10pt,twocolumn,letterpaper]{article}

\usepackage[pagenumbers]{cvpr} 

\usepackage{makecell}
\usepackage{bbding}
\usepackage{multirow}

\definecolor{cvprblue}{rgb}{0.21,0.49,0.74}
\usepackage[pagebackref,breaklinks,colorlinks,allcolors=cvprblue]{hyperref}

\usepackage{colortbl}

\def\paperID{12834} 
\def\confName{CVPR}
\def\confYear{2025}

\title{InfiniteWorld: A Unified Scalable Simulation Framework for \\ General Visual-Language Robot Interaction}
\author{
Pengzhen Ren\textsuperscript{1}\footnotemark[1] \quad Min Li\textsuperscript{2}\footnotemark[1] \quad Zhen Luo\textsuperscript{3}\footnotemark[1] \quad Xinshuai Song\textsuperscript{2}\footnotemark[1]\quad Ziwei Chen\textsuperscript{3}\footnotemark[1] \quad Weijia Liufu\textsuperscript{2}\footnotemark[1] \\
Yixuan Yang\textsuperscript{3}\footnotemark[1] \quad Hao Zheng\textsuperscript{3}\footnotemark[1] \quad Rongtao Xu\textsuperscript{4} \quad Zitong Huang\textsuperscript{3} \quad Tongsheng Ding\textsuperscript{3}\quad Luyang Xie\textsuperscript{3}\\
\quad Kaidong Zhang\textsuperscript{2} \quad Changfei Fu\textsuperscript{3} \quad Yang Liu\textsuperscript{2} \quad Liang Lin\textsuperscript{2} \quad Feng Zheng\textsuperscript{3}\footnotemark[2] \quad Xiaodan Liang \textsuperscript{2,4}\footnotemark[2]\\
{\normalsize
\textsuperscript{1}Peng Cheng Laboratory \quad       
\textsuperscript{2}Sun Yat-sen University \quad
\textsuperscript{3}Southern University of Science and Technology \quad
\textsuperscript{4}MBZUAI
}\\
\tt\small pzhren@foxmail.com, \{linm57, songxsh, liufwj5, zhangkd3\}@mail2.sysu.edu.cn, \\
\tt\small luoz2024@mail.sustech.edu.cn, \{liuy856, liangxd9\}@mail.sysu.edu.cn, \\
\tt\small xurongtao2019@ia.ac.cn, \{linliang, f.zheng\}@ieee.org
\\
\normalsize \color{red}{https://github.com/pzhren/InfiniteWorld}
}

\begin{document}
\maketitle
\renewcommand{\thefootnote}{\fnsymbol{footnote}} 
\footnotetext[1]{Equal contribution} 
\footnotetext[2]{Corresponding authors} 
\begin{abstract}
Realizing scaling laws in embodied AI has become a focus. However, previous work has been scattered across diverse simulation platforms, with assets and models lacking unified interfaces, which has led to inefficiencies in research.
To address this, we introduce InfiniteWorld, a unified and scalable simulator for general vision-language robot interaction built on Nvidia Isaac Sim.
InfiniteWorld encompasses a comprehensive set of physics asset construction methods and generalized free robot interaction benchmarks.
Specifically, we first built a unified and scalable simulation framework for embodied learning that integrates a series of improvements in generation-driven 3D asset construction, Real2Sim, automated annotation framework, and unified 3D asset processing.
This framework provides a unified and scalable platform for robot interaction and learning.
In addition, to simulate realistic robot interaction, we build four new general benchmarks, including scene graph collaborative exploration and open-world social mobile manipulation.
The former is often overlooked as an important task for robots to explore the environment and build scene knowledge, while the latter simulates robot interaction tasks with different levels of knowledge agents based on the former.
They can more comprehensively evaluate the embodied agent's capabilities in environmental understanding, task planning and execution, and intelligent interaction. 
We hope that this work can provide the community with a systematic asset interface, alleviate the dilemma of the lack of high-quality assets, and provide a more comprehensive evaluation of robot interactions.

\end{abstract}

\section{Introduction}
\label{sec:intro}

Building an infinite world for embodied artificial intelligence (AI) \cite{duan2022embodiedAIsurvey} that allows robots to interact and learn freely in an open environment like humans is an important direction of the embodiment community.
To achieve this, the robotic simulation learning platform must possess several critical attributes: fast and precise physical simulation, user-friendly and expeditious interface design, highly realistic and varied 3D assets, and a comprehensive robot interactive task design.
Recently, NVIDIA's Omniverse Isaac Sim \cite{IsaacSim} has achieved excellent results in physically based rendering, low-level interaction complexity, deformation simulation, \etc.
However, previous work \cite{wang2024grutopia, mittal2023orbit, yang2024physcene, wu2024metaurban, savva2019habitat} still lacked a systematic and unified design in asset construction and interaction design, resulting in fragmented efforts and repetitive tasks within the community.
Therefore, considering how to achieve \textbf{scaling laws} and \textbf{realistic robot interaction} in the field of embodied AI has become two major issues of concern in the industry.

Recent advancements in AI, particularly in multimodal large-scale language models (MLLM) \cite{liu2024improvedLLaVa, touvron2023llama, achiam2023gpt4}, have been propelled by vast Internet-scale data. In contrast, robotics data remains sparse compared to the abundant visual and linguistic resources online.
A straightforward approach is to collect large-scale robot data directly in the real world like Open X-Embodiment \cite{o2023openxembodiment} and DROID \cite{khazatsky2024droid}.
However, they are severely limited by high data collection costs and generalization issues across different hardware platforms.
Therefore, simulation is presented as a promising alternative.
In order to implement the scaling laws of embodied AI, the community has made a lot of attempts.
For example, previous work \cite{gao2024embodiedcity, nasiriany2024robocasa, wang2024grutopia, wu2024metaurban, liu2023aerialvln} used AI generation tools \cite{nasiriany2024robocasa, tochilkin2024triposr, yang2024physcene, liu2024cage}, or semi-automated \cite{nasiriany2024robocasa, gao2024embodiedcity} or manual design methods \cite{wang2024grutopia} to build 3D scene and object assets.
The creation or collection of these high-quality assets is labor-intensive and often fragmented across various simulation platforms, hindering their efficient use.
We believe that this dilemma mainly stems from the lack of a unified and high-quality embodied asset construction interface in current simulation platforms.

On the other hand, previous embodied benchmarks predominantly focus on conventional tasks like object localization, navigation, or manipulation. Recently, there's a growing interest in social navigation \cite{puig2023habitat3, wang2024grutopia}, which more closely resembles human interaction.
In particular, GRUtopia \cite{wang2024grutopia} proposes a non-player character (NPC) with a global view and uses it as an interactive object in the robot navigation task to assist it in completing corresponding ambiguous tasks.
However, it is constrained by the lack of a character with a “God's perspective” in reality, which limits its ability to fully simulate real-world interactions. 
Especially in special scenarios where communication is limited (such as coal mines), this requires robots to have the ability to explore independently and complete tasks collaboratively.
We believe that simulating more realistic human interactions is crucial to assess the capabilities of embodied agents at the levels of task reasoning, planning, perception, and interaction, but current interactions in simulators still have significant gaps from the real world.


Based on the above observations, in this work, we aim to build an infinite world of unified robot interaction simulation platforms based on the NVIDIA Isaac Sim: comprehensive physical asset construction and universal free robot interaction.
For assets, we have designed multiple asset interfaces for the InfiniteWorld simulation to enable unlimited scaling of scene and object assets.
Specifically, we first integrated a generation-driven 3D asset construction method for Isaac Sim, which includes: language-driven 3D scene generation, controllable joint object generation, and image-to-3D object reconstruction. 
Among them, our language-driven 3D scene reconstruction method built based on HOLODECK \cite{yang2024holodeck} can achieve 200+ different scene style changes, as well as various object edits (\eg, color/texture/quantity/replacement/removal/addition, \etc.).
This can help us easily achieve infinite expansion of the scene.
We also build a Real2Sim pipeline based on the improved PGSR \cite{chen2024pgsr}, which covers the entire process from photographic data to accurate and visually coherent models.
Additionally, we establish an automated annotation platform Annot-8-3D with optional AI-assisted human-in-the-loop capabilities. It supports distributed collaboration and producing comprehensive annotation data, which streamlines the creation of scene assets and the formulation of interactive tasks.
Finally, we've unified various open-source scenes (\eg, HSSD \cite{10657917hssd}, HM3D \cite{HM3D}) and object assets (\eg, 3D Front \cite{3DFront}, PartNet-mobility \cite{mo2019partnet}) onto the Isaac Sim platform,  greatly enhancing asset utilization.
As a unified and extensible simulation framework, InfiniteWorld can provide the community with rich and massive embodied assets and accelerate the arrival of embodied scaling laws.

For interaction, which is the core of robot activities, how to simulate more realistic human-like interactions is of great significance for evaluating the agent's environment perception, task understanding, planning, and execution in the open world. Among them, social interaction is the key to human-robot interaction.
To enhance the realism of robot interactions, InfiniteWorld introduces two novel tasks beyond traditional navigation and manipulation: \textit{(i) Scene Graph Collaborative Exploration (SGCE)} and \textit{(ii) Open-World Social Mobile Manipulation (OWSMM)}.
First, similar to how humans observe and build world knowledge, robots construct scene graphs about the environment, which is the most important step in perceiving and understanding the environment. 
However, previous benchmarks \cite{wang2024grutopia, wu2024metaurban} often ignore this point when constructing tasks. 
To address this, we developed the SGCE task to assess an agent's capability in building environmental knowledge through free exploration and collaboration, thereby equipping them to handle more complex interactive tasks.
Furthermore, social interaction is the key to human interaction. In order to simulate more realistic human interaction, we designed two levels of interaction tasks for OWSMM based on SGCE: hierarchical interaction and horizontal interaction.
Specifically, hierarchical interaction simulates social mobile manipulation with an “administrator” environment. The administrator has more complete environmental knowledge than ordinary agents and provides question-and-answer services for ordinary agents when performing ambiguous and complex tasks to assist the agents in completing tasks.
Horizontal interaction requires that all agents have the ability to obtain scene knowledge equally, and they can exchange scene knowledge through social interaction to complete tasks together.

Our main contributions are as follows:
\begin{itemize}
    \item We have built a unified and scalable simulation framework that integrates various improved and latest embodied asset reconstruction methods.
    This has greatly alleviated the community's plight of lacking high-quality embodied assets.
    \item We build a complete web-based smart point cloud automatic annotation framework that supports distributed collaboration, AI assistance, and optional human-in-the-loop features. This provides strong support for complex robot interactions.
    \item Finally, we designed systematic benchmarks for robot interaction, including scene graph collaborative exploration and open-world social mobile manipulation. This provides a comprehensive and systematic evaluation of the capabilities of embodied agents in perception, planning, execution, and communication.

\end{itemize}

\section{Related Work}
\label{sec:related_work}

\begin{table*}[t]
\resizebox{1\textwidth}{!}{
\centering
\begin{tabular}{l|ccc|c|ccc|ccccc}
\toprule
\multirow{3}{*}{Name}& \multicolumn{3}{c|}{Asset}&  \multirow{3}{*}{\makecell{Annotation\\Platform}}  &\multicolumn{3}{c|}{Robotic Platforms}             & \multicolumn{5}{c}{Benchmark}      \\ 
& \multirow{2}{*}{\makecell{Scene\\Authoring}} & \multirow{2}{*}{Object} & \multirow{2}{*}{\makecell{Unified\\Asset}} &&\multirow{2}{*}{Fixed-M}&\multirow{2}{*}{Mobile-M}&\multirow{2}{*}{Legged}& \multirow{2}{*}{\makecell{Language\\Instruction}} & \multirow{2}{*}{\makecell{Scene Graph \\Exploration}} & \multicolumn{2}{c}{Social Interaction} & \multirow{2}{*}{Action} \\
&                 &        &         & &&       &                                       &                                  &        & Hierarchical        & Horizontal       &        \\\midrule
 Maniskill2 \cite{ehsani2021manipulathor}&- &- &\XSolidBrush&\XSolidBrush &\Checkmark & \Checkmark&\XSolidBrush &\Checkmark &\XSolidBrush & \XSolidBrush&\XSolidBrush&\textit{M}\\
 Social Navigation \cite{puig2023habitat3}&\textit{M} & -&\XSolidBrush &\XSolidBrush&\XSolidBrush & \Checkmark &\Checkmark &\Checkmark&\XSolidBrush&\XSolidBrush&\Checkmark&\textit{N,M}\\
 HomeRobot \cite{yenamandra2023homerobot} &\textit{M} &- &\XSolidBrush &\XSolidBrush& \XSolidBrush&\Checkmark &\XSolidBrush &\Checkmark &\XSolidBrush&\XSolidBrush&\XSolidBrush&\textit{N,M}\\
 VLN-CE \cite{krantz2020beyond}&\textit{M} &\textit{I} &\XSolidBrush &\XSolidBrush&\XSolidBrush &\Checkmark &\XSolidBrush &\Checkmark &\XSolidBrush&\XSolidBrush&\XSolidBrush&\textit{N}\\
 ProcTHOR-10k \cite{deitke2022️}&\textit{P,M} &\textit{P} &\XSolidBrush &\XSolidBrush&\XSolidBrush &\Checkmark &\XSolidBrush &\XSolidBrush &\XSolidBrush&\XSolidBrush&\XSolidBrush&\textit{N,M}\\
 ManipulaTHOR \cite{ehsani2021manipulathor}&-&-&\XSolidBrush &\XSolidBrush&\XSolidBrush &\Checkmark &\XSolidBrush &\XSolidBrush &\XSolidBrush&\XSolidBrush&\XSolidBrush&\textit{N}\\
 ALFRED \cite{shridhar2020alfred}&- &- &\XSolidBrush&\XSolidBrush &\XSolidBrush &\Checkmark &\XSolidBrush &\Checkmark &\XSolidBrush&\XSolidBrush&\XSolidBrush&\textit{N,M}\\
 Arnold \cite{gong2023arnold}&\textit{M} &-&\XSolidBrush &\XSolidBrush&\Checkmark &\XSolidBrush &\XSolidBrush &\Checkmark &\XSolidBrush&\XSolidBrush&\XSolidBrush&\textit{M}\\
 Behavior-1K \cite{li2023behavior-1k}&\textit{P,M} &- &\XSolidBrush&\XSolidBrush &\XSolidBrush &\Checkmark &\XSolidBrush &\Checkmark &\XSolidBrush&\XSolidBrush&\XSolidBrush&\textit{N,M}\\
 Orbit \cite{mittal2023orbit}&\textit{M} &- &\XSolidBrush &\XSolidBrush&\Checkmark &\Checkmark &\Checkmark &\XSolidBrush &\XSolidBrush&\XSolidBrush&\XSolidBrush&\textit{N,M}\\
 GRUtopia \cite{wang2024grutopia}&- &- &\XSolidBrush &\Checkmark&\Checkmark &\Checkmark & \Checkmark& \Checkmark&\XSolidBrush&\Checkmark&\XSolidBrush&\textit{N,M}\\\midrule
InfinitedWorld &\textit{P,M,E           }& \textit{P,I }   & \Checkmark  & \Checkmark & \Checkmark             & \Checkmark                                     & \Checkmark         & \Checkmark                                     & \Checkmark                                        & \Checkmark                   & \Checkmark                & \textit{N,M}    \\  
\bottomrule
\end{tabular}
}

\caption{Comparison of InfinitedWorld with other platforms in terms of assets, robotic platforms, and benchmarks. In the \textbf{Asset} column, \textit{P} stands for unlimited programmatic automatic generation, \textit{M} stands for mesh-scan scenes, \textit{E} stands for language-driven scene editing, and \textit{I} for image-based object generation. In the \textbf{Social Interaction} column, Hierarchical and Horizontal represent social interactions with and without administrators, respectively. In the \textbf{Action} column, \textit{N} and \textit{M} stand for navigation and manipulation, respectively. 
“-” indicates that it is not applicable or has no relevant function.
}
\vspace{-1em}
\label{tab:simulator}
\end{table*}

\subsection{Embodied AI Simulators}
Currently, many simulators have been developed for embodied AI-related research \cite{savva2019habitat, puig2023habitat3, xia2018gibson, wani2020multion, ramakrishnan2021habitat_hm3d, deitke2020robothor, chen2020soundspaces, james2020rlbench, deitke2022️, mittal2023orbit, yang2024physcene, wu2024metaurban}. 
They mainly focuses on the improvement of realistic physical simulation and the diversity of task design. For example, in physics simulation, from abstracting physical interactions into symbolic reasoning (\textit{eg.}, VirtualHome \cite{puig2018virtualhome} and Alfred \cite{shridhar2020alfred}) to conducting navigation research in 3D scanning scenes (\textit{eg.}, Habitat \cite{savva2019habitat}), to realistic actions, environmental interactions and physical simulations (\textit{eg.}, Habitat 2.0 \cite{szot2021habitat2}, ManiSkills \cite{gu2023maniskill2}, TDW \cite{gan2022threedworld_TDW}, SoftGym \cite{lin2021softgym}, RFUniverse \cite{fu2022rfuniverse} and iGibson \cite{shen2021igibson, li2021igibson2.0}), the gap between virtual and real environments are gradually being narrowed.
In terms of task design, current work mainly explores the diversity of embodied AI task settings \cite{zeng2021transporter, james2020rlbench, mees2022calvin, ren2024surferprogressivereasoningworld}. 
For example, RoboGen \cite{wang2023robogen} and MimicGen \cite{mandlekar2023mimicgen} use generative models and LLM to generate tasks, Surfer \cite{ren2024surferprogressivereasoningworld} and HandMeThat \cite{wan2022handmethat} study hierarchical reasoning tasks for desktop manipulation, GRUtopia \cite{wang2024grutopia} and Habitat 3.0 \cite{puig2023habitat3} study social interaction, etc.
Different from the above work, we aim to build an infinite world of embodied AI based on Isaac Sim: it has infinite scene and object assets driven by generation, human-like open-world social interaction, realistic physics simulation, and unified 3D assets. 
This will provide the community with strong support for realizing the scaling of embodied AI.
The detailed comparison between InfiniteWorld simulation and other platforms is presented in Table \ref{tab:simulator}.

\subsection{Interaction in Simulator}
Social interaction in embodied AI is the interaction method closest to humans and is also the key to human-robot interaction research. 
For example, Habitat 3.0 \cite{puig2023habitat3} proposes a human-in-the-loop paradigm that uses LLMs to simulate authentic human behaviors to explore collaboration between humanoid and robotic agents in home environments. 
Furthermore, GRUtopia \cite{wang2024grutopia} designed a NPC with global ground truth environment information. It is used for human-robot interaction, providing key interactive information to robots, helping robots complete complex tasks, and simulating real-world social interactions.
This NPC design goes beyond the traditional human-in-the-loop paradigm to a certain extent, but there is no NPC with global environmental information in the real world, which is detrimental to simulating real social interaction.
To this end, this study proposes an LLM-driven human-like interaction paradigm based on environment exploration to simulate real human interaction.

\subsection{Scene and Asset Handling}
Scaling of the implemented simulation platform assets is one of the most critical issues in the current development of embodied AI, and it is the basis for obtaining large-scale robot datasets. 
To this end, the community has studied various embodied asset generation technologies, such as realistic virtualization of real scenes based on 3D Gaussian splatter technology \cite{kerbl20233dgs, chen2024pgsr}, large-scale 3D scenes \cite{yang2024physcene, yang2024holodeck} and 3D objects \cite{tochilkin2024triposr}, and articulated object asset generation.
However, they often lack a unified and effective interface and cannot be fully applied. We built a unified interface for them based on the Isaac Sim platform and achieved unlimited expansion of 3D assets.
\section{InfiniteWorld Simulation}
\label{sec:Simulator}
In this section, we will focus on the InfiniteWorld simulator's approach to building large-scale assets. Specifically, our simulator supports generative AI-driven 3D asset reconstruction, improved Real2Sim scene reconstruction, a web-based smart point cloud automatic annotation framework Annot8-3D, and a unified 3D asset library.

\subsection{Generate-Driven 3D Asset Construction}

\begin{figure*}[th]
    \centering
    \includegraphics[width=0.8\linewidth]{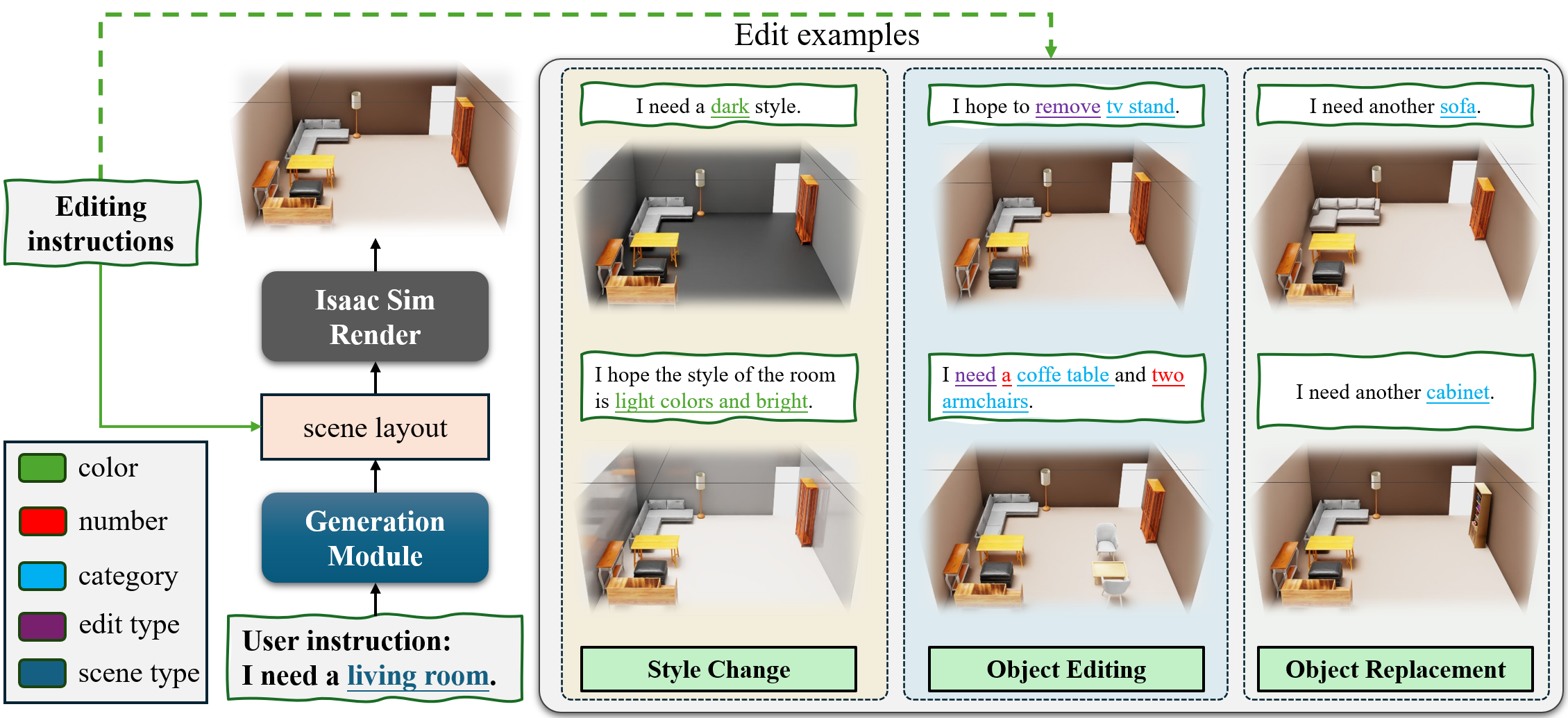}
    \caption{Language-driven automatic scene generation and editing framework based on HOLODECK \cite{yang2024holodeck}. It can easily generate various interactive high-fidelity scenes that meet the requirements of users, including scene style replacement, object editing (\eg, adding/removing a specific number of objects), and replacement (that is, replacing similar objects), \etc.
    }
    \label{fig:text2scene}
\end{figure*}

Building a large-scale, interactive, realistic environment for a simulator platform is critical for embodied learning.
Cost and diversity are the main limitations plaguing the construction of large-scale 3D environments.
Leveraging language as a driver for large-scale scene generation \cite{lin2024instructscene,yang2024llplace} is a popular solution.
In particular, HOLODECK \cite{yang2024holodeck} can use text as a driver and leverage a widespread 3D asset database to create 3D environments with accurate semantics, good spatial layout, and interactivity.
In addition, inspired by the use of hand-designed scene styles to expand scene assets in RoboCasa \cite{nasiriany2024robocasa}, we implemented automated expansion of large-scale user-defined scene assets on Isaac Sim based on HOLODECK \cite{yang2024holodeck}. It supports free replacement of 236 different textures of floors, and walls.
This means that the number of our scenes can be easily expanded 236 times.
As well as editing operations such as similar replacement, deletion, addition, and texture replacement of object assets in the scene.
This provides a unified and efficient interface for large-scale automated scene generation.
Based on the above method, we first constructed 10K indoor scenes, mainly including household and social environments.
For household scenes, we simulated the layout of real household scenes and generated 1-5 different room numbers for each scene to meet different task requirements.
Social environments include many scenes such as offices, restaurants, bars, gyms, and shops.
And, we also used scene-style replacement to generate a total of 2.36 M scenes. 
Figure \ref{fig:text2scene} shows some examples of language-driven automated scene generation and editing.
These constructed scenarios will be published upon acceptance of the paper.

In addition, we have integrated single image to 3D object asset reconstruction \cite{tochilkin2024triposr} and controllable articulation generation \cite{liu2024cage} in the InfiniteWorld simulator to further enrich our asset library. This provides a large number of diverse interactive scenarios for embodied agent learning.

\subsection{Depth-Prior-Constrained Real2Sim}
Recently, 3D Gaussian Splatting (3DGS) \cite{kerbl3Dgaussians} variants represented by GauStudio \cite{ye2024gaustudio}, SuGaR \cite{guedon2024sugar}, and PGSR \cite{chen2024pgsr} have achieved high-quality mesh reconstruction effects while providing explicit geometry information.
However, they have difficulty resolving the complexities created by reflections on smooth surfaces. If these reflections are not handled correctly, they can significantly interfere with the foundational steps of point cloud initialization, specifically during the structure-from-motion (SfM) phase.
To alleviate these issues, we introduce two types of regularization loss based on depth and normal vector based on PGSR \cite{chen2024pgsr}.
Specifically, we employ a pre-trained depth estimation model, Depth Pro \cite{bochkovskii2024depth}, to generate depth estimates within the camera coordinate system for each RGB image. Additionally, we use the Local Plane Assumption \cite{chen2024pgsr} from PGSR to compute plane normal vectors, thereby providing extra supervision for the single-view loss in PGSR.

\begin{figure*}
    \centering
    \includegraphics[width=0.8\linewidth]{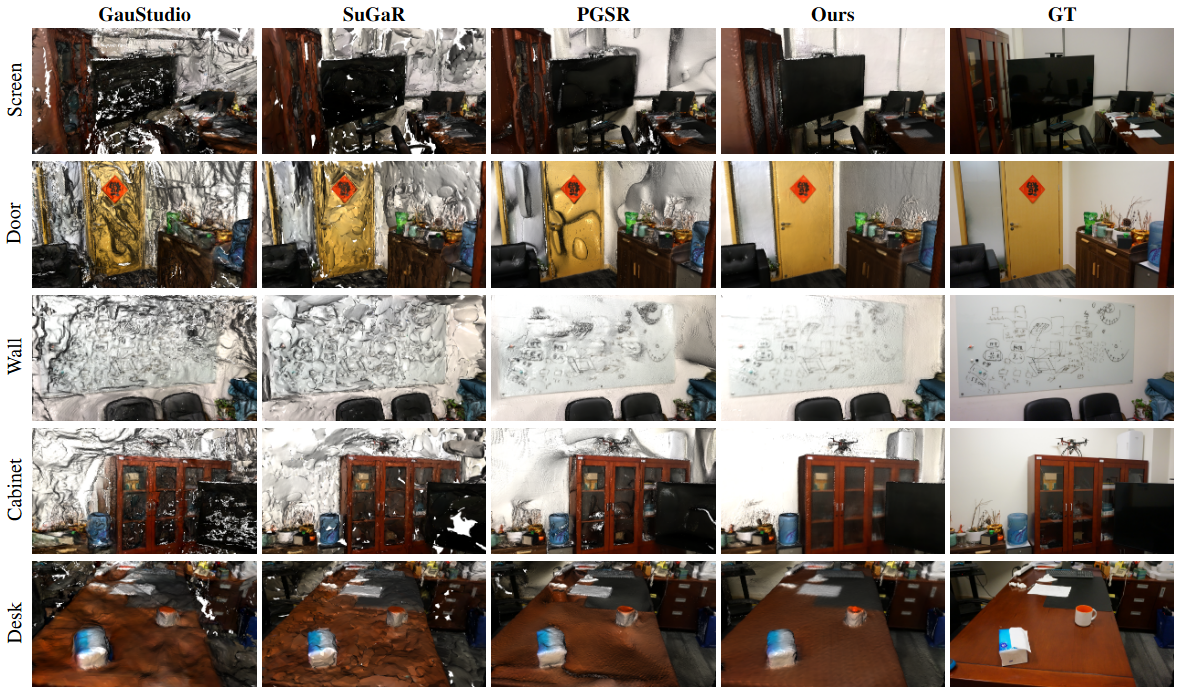}
    \caption{The reconstruction visual comparisons on test-set views among GauStudio, SuGaR, PGSR, and our proposed method from real-world captured images of an office. Compared to 3DGS and SuGaR, PGSR provides an improved visual experience. Building upon PGSR, our method incorporates regularization loss terms for depth and normal vectors, achieving smoother planar surfaces, such as walls, doors, and screens, and demonstrating more robust handling of transparent surfaces like glass.}
    \label{fig:compare_of_reconstruct}
\end{figure*}

Figure~\ref{fig:compare_of_reconstruct} shows a comparison of the reconstruction effects of related methods in a real office scene. As shown in Figure \ref{fig:compare_of_reconstruct}, PGSR \cite{chen2024pgsr} produced the highest quality meshes in our scene reconstruction task.
In contrast, our improved method is able to generate refined meshes when dealing with certain planar and reflective surfaces.
In addition, we also designed a complete post-processing step for the reconstructed scene to further optimize the model with respect to issues such as axis alignment, noise, surface continuity, and size.
More details about our Real2Sim pipeline and the comparison of the results before and after post-processing are shown in the Appendix \ref{app:Real2SimPipeline}.


\begin{figure}[t]
    \centering
    \includegraphics[width=1\linewidth]{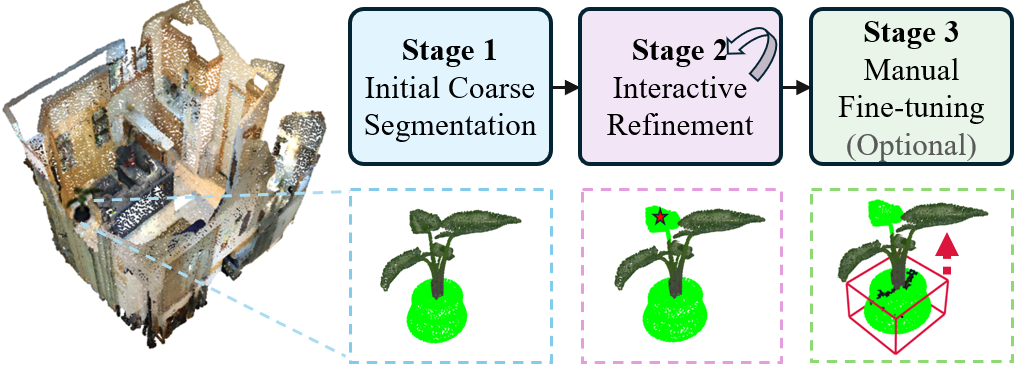}
    \caption{The Annot8-3D framework pipeline. 
    }
    \label{fig:annot8-3d}
    \vspace{-1em}
\end{figure}

\begin{figure*}
    \centering
    \includegraphics[width=0.85\linewidth]{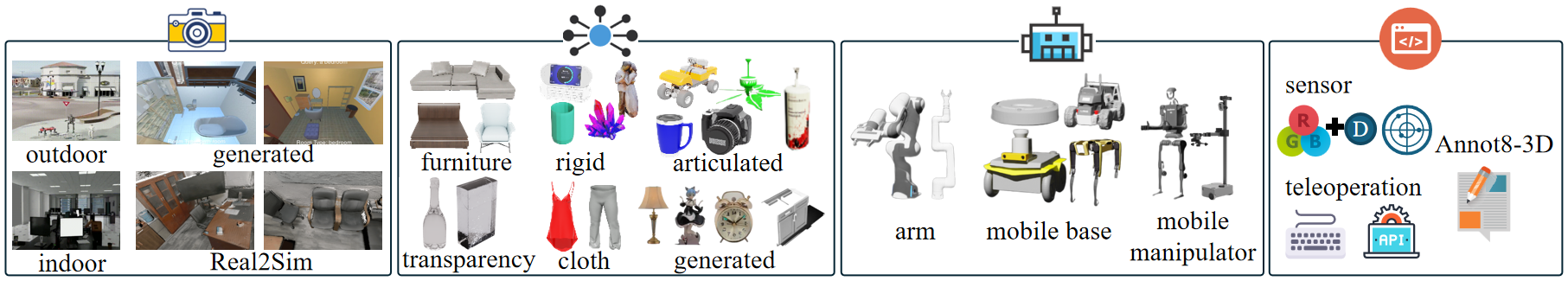}
    \caption{Overview of the functions of InfiniteWorld simulator.
    Our simulation platform supports different sensors, robot platforms, and teleoperation. In addition, it also realizes unlimited expansion of scene and object assets through generative and Sim2Real methods, and we have also built an annotation platform to reduce annotation costs and improve annotation quality.
    }
    \label{fig:simulator}
    \vspace{-1em}
\end{figure*}

\subsection{Annot8-3D: Automatic Annotation Framework}
We also proposed Annot8-3D, a novel web-based smart point cloud automatic annotation framework that combines AI-assisted automation with human-in-the-loop refinement for efficient and accurate 3D point cloud labeling. The framework implements a multi-stage annotation pipeline that progressively refines segmentation results through coarse-to-fine labeling, leveraging state-of-the-art deep learning models while allowing human guidance when needed. 
Specifically, figure \ref{fig:annot8-3d} shows the multi-stage annotation pipeline of Annot8-3D, which mainly contains three stages: initial coarse segmentation, interactive refinement, and manual fine-tuning. First, in the initial coarse segmentation stage, the pipeline begins with automated coarse-grained segmentation using Point Transformer V3 \cite{wu2024ptv3}, which provides initial object proposals across the point cloud. Second, in the interactive refinement stage, the system enables human reviewers to examine and refine the coarse segmentation results through positive and negative prompts that guide focused refinement of specific regions. This stage integrates SAM2Point \cite{guo2024sam2point} to process these prompts and generate refined segmentations, allowing for iterative refinement loops until satisfactory results are achieved. Finally, for cases where automated refinement proves insufficient, the manual Fine-tuning stage provides manual segmentation tools for precise adjustments.
A detailed feature comparison between Annot8-3D and existing annotation tools is provided in Appendix \ref{appendix:annot8-3d}.

\subsection{Unified 3D Asset}
In addition, we have also integrated some open-source 3D assets into the InfiniteWorld simulator. 
Currently, existing popular 3D assets often have different simulation platforms and different data formats.
The lack of a unified data format between different simulation platforms makes asset interoperability difficult.
To this end, we provide a unified interface for assets from different simulation platforms based on Isaac Sim. All assets are unified into \textit{.usd}, thus realizing the unified calling of different assets on the Isaac Sim platform.
Specifically, we provide conversion scripts from different formats to usable formats to facilitate physical simulation in Isaac Sim.
It includes 3D scene-level assets (\eg HSSD \cite{10657917hssd}, HM3D\cite{HM3D}, Replica\cite{straub2019replica} and Scannet \cite{dai2017scannet}) and 3D object-level assets (\eg \textit{3D Front} \cite{3DFront}, \textit{PartNet-mobility} \cite{mo2019partnet}, \textit{Objaverse (Holodeck)} \cite{deitke2024objaverse}, and \textit{ClothesNet} \cite{zhou2023clothesnet}).

The unified object assets cover a wide range of categories such as fruits, beverages, dolls, appliances, furniture, etc. It also includes some commonly used articulated objects. In addition, on the Isaac Sim platform, we have also implemented the simulation of special objects such as soft bodies and transparency. This is beneficial for achieving realistic physics simulations in a simulation environment. This provides strong support for embodied agents to perform various complex manipulation tasks. 
Overall, the processed unified 3D asset statistics are shown in Appendix \ref{app:asset}.

Overall, the main features of the InfiniteWorld simulator are shown in Figure \ref{fig:simulator}.

\section{Experiments}
\label{sec:experiments}

\begin{figure*}
    \centering
    \includegraphics[width=0.84\linewidth]{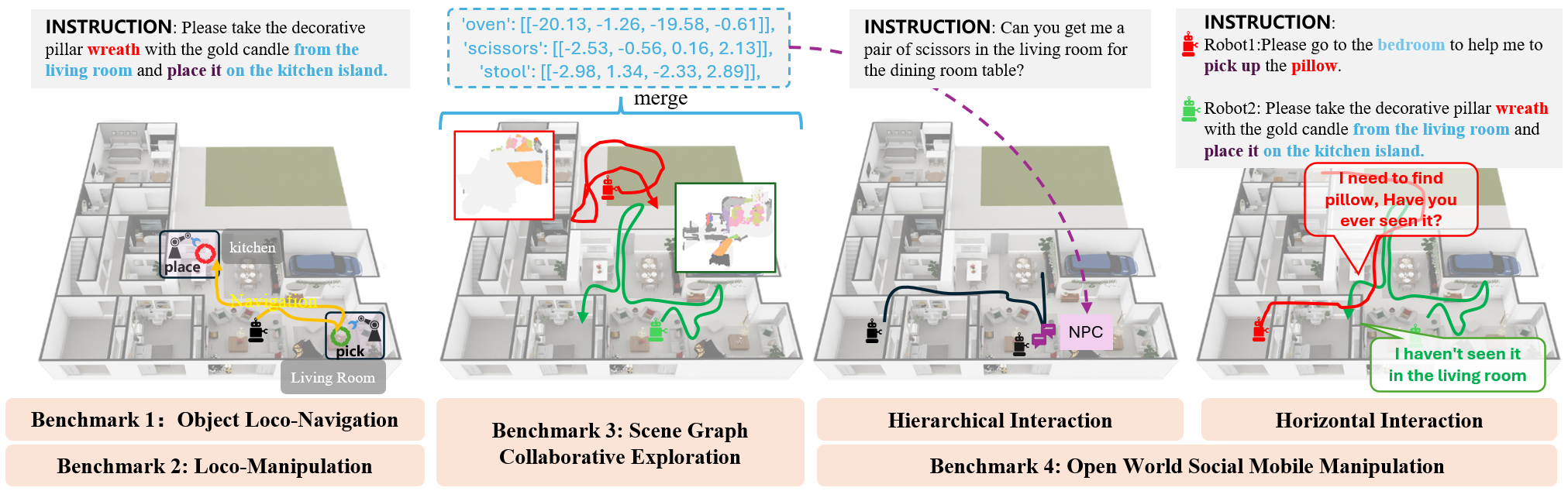}
    \vspace{-0.5em}
    \caption{An overview of the proposed benchmark.}
    \label{fig:benchmark}
    \vspace{-1em}
\end{figure*}
\subsection{Benchmark}
\textbf{Benchmark 1: Object Loco-Navigation.} The object loco-navigation task evaluates the agent's basic ability to navigate to the target object given language instructions. The task succeeds if the target object appears in the agent's field of view.
The agent needs to search for and locate specific objects in specific areas within the scene. 
When the distance between the robot and the target object is less than 2 meters and the object is within 60 degrees of the robot's horizontal field of view, the task execution is successful.

\noindent \textbf{Benchmark 2: Loco-Manipulation.} 
Based on the object loco-navigation task, we developed a loco-manipulation task. This task validates the agent's basic ability in navigation, manipulation, and planning. The agent needs to understand natural language instructions, locate the correct object, perform the appropriate actions to move the object to the target position, and finally successfully place it down.


\noindent \textbf{Benchmark 3: Scene Graph Collaborative Exploration.}
In traditional single-robot systems, the robot explores unknown areas sequentially, gradually building up the scene graph. However, this approach is often inefficient in large-scale or dynamically changing environments, limiting the speed of scene graph construction and the richness of information obtained. Introducing multi-agent scene graph construction can significantly improve the efficiency and quality of this process. Multiple robots work collaboratively, sharing information and merging their views to build a unified scene graph. While each robot independently perceives and maps parts of the environment, the agents share map data, update object semantic labels, and synchronize their positions via wireless communication, effectively boosting mapping efficiency.

\noindent \textbf{Benchmark 4: Open World Social Mobile Manipulation.} 
In this benchmark, we designed an open-world social mobile manipulation. 
It mainly includes two interaction methods: hierarchical interaction and horizontal interaction. 
The former simulates embodied AI interaction with hierarchical knowledge structure, and the latter simulates embodied AI interaction with equal knowledge acquisition capabilities.
\begin{itemize}
    \item \textbf{Hierarchical interaction.} In hierarchical interaction tasks, it is used to simulate the agent interaction mode with a hierarchical knowledge structure in the environment. For example, compared to ordinary agents, administrators (such as salespersons, etc.) clearly have more knowledge about the environment. Encouraging agents to have conversations with administrators, can help agents better understand user intentions and improve task execution success rates. Specifically, we use the scene graph explored in benchmark 3 to construct an administrator role with high-level knowledge, and the agent is required to ask the administrator questions as much as possible to complete the instruction tasks accurately and efficiently.
    \item \textbf{Horizontal interaction.} In horizontal interaction tasks, it is used to simulate the “passer-by interaction scene”. There is no administrator with a “God's perspective” in the scene, and all agents can obtain scene knowledge equally. Specifically, the scene contains multiple agents with the same status. They can independently build their own scene graphs and transfer knowledge through social dialogue to improve the efficiency and success rate of task completion.
\end{itemize}

Some more detailed benchmark settings such as instruction format, task settings, \etc. are shown in Appendix \ref{app:benchmarkSetting}.

\vspace{-0.5em}

\subsection{Settings}
\begin{itemize}
    \item \textbf{Robot Setups.} We use the Stretch robot as the execution agent for all experiments. It has a mobile base with omnidirectional wheels and a 7-degree-of-freedom (DOF) manipulator, allowing it to effectively perform mobile manipulation tasks.
    \item \textbf{Task Generation.} We use GPT-4o \cite{achiam2023gpt4o} and combine the scene semantics of the HSSD \cite{10657917hssd} dataset to generate corresponding task instructions.
\end{itemize}
We also provide a variety of interfaces for different levels of tasks. More details about the occupancy map, path planning, and manipulation settings are shown in Appendix \ref{app:TaskSettingDetails}.

\subsection{Baselines}
\begin{itemize}
    \item \textbf{LLM-Based Instruction Following.}
Based on a large language model (LLM) and prompt engineering, we decompose natural language instructions into action interfaces that can be executed by embodied agent, guiding it step by step to complete tasks.
    \item  \textbf{VLM Zero-Shot.}
By inputting the global scene information and current observations into a vision-language model (VLM), we use prompt engineering to output the actions that the agent should execute. 
\end{itemize}

\begin{itemize}
\item \textbf{Single Semantic Map.}
We use the method proposed in Goal-Oriented Semantic Exploration \cite{chaplot2020object} for 2D semantic mapping, while employing the FBE \cite{fbe} algorithm as the global planner in combination with the FMM \cite{fmm} planning algorithm for local planning.

\item \textbf{Random.} In the robot's action space, actions are randomly sampled for execution, or target points are randomly sampled in the planning space, and planning algorithms are used to solve for them.

\item  \textbf{LLM-Based Planning.}
Using the Co-NavGPT \cite{yu2023Co-NavGPT}, we employ a large language model (LLM) as a planner for multi-agent systems. The merged observation map of the agents is converted into a textual description, which is then processed by the LLM to perform goal planning for multiple agents.

\item  \textbf{LLM-Planner \cite{song2023llmplanner}} is a few-shot grounded planning model. Different from common planning models, LLM-Planner uses LLMs to generate plans directly instead of ranking acceptable skills, reducing the need for sufficient prior knowledge of the environment and the number of calls to LLMs. Re-planning of LLM-Planner allows it to dynamically adjust the planning based on current observations, resulting in more informed plans.
\end{itemize}
We also evaluated the capabilities of different LLMs (\eg, GPT-4o \cite{achiam2023gpt4o}, Qwen-turbo, and Chat-GLM4-flash) and VLMs (\eg, GPT-4o \cite{achiam2023gpt4o}, Qwen-VL2, and GLM-4v) in task planning and scene perception.


\begin{table}[]
    \centering
    \small
    \begin{tabular}{l|c|c|c|c}
    \toprule
    Method & LLM/VLM &SR & SPL & NE \\
    \midrule
    \makecell{LLM-Based Ins \\Following} &\makecell{GPT-4o\\Qwen-turbo\\Chat-GLM4-flash}& \makecell{90.82\\69.94\\66.41} & \makecell{90.82\\69.94\\66.41} & \makecell{1.00\\1.00\\0.96} \\\hline
    VLM Zero-Shot &\makecell{GPT-4o\\Qwen-VL2\\GLM-4V}&\makecell{0.06\\0.00\\0.00} &\makecell{0.00\\0.00\\0.00} &\makecell{15.23\\11.67\\26.53}\\
    \bottomrule
    \end{tabular}
    \caption{Object Loco-Navigation}
    \label{tab:obj-loco}
\end{table}


\begin{table}[]
    \centering
    \small
    \begin{tabular}{l|c|c|c|c}
    \toprule
        Method &LLM/VLM& SR & SPL & NE \\
    \midrule
        \makecell{LLM-Based \\Ins Following} &\makecell{GPT-4o\\Qwen-turbo\\Chat-GLM4-flash}& \makecell{77.28\\42.64\\50.63} & \makecell{77.28\\42.64\\50.63} & \makecell{0.94\\0.93\\0.93} \\\hline
        VLM Zero-Shot &\makecell{GPT-4o\\Qwen-VL2\\GLM-4V}& \makecell{0.01\\0.00\\0.00}&\makecell{0.00\\0.00\\0.00}& \makecell{15.37\\12.05\\26.50}\\
    \bottomrule
    \end{tabular}
    \caption{Loco-Manipulation}
    \label{tab:loco-maniplation}
\end{table}


\begin{table}[]
    \centering
    \small
    \begin{tabular}{l|c|c|c}
    \toprule
        Method &VLM& SER & MRMSE \\
    \midrule
        Single SemMap &-& 0.2581 & 5.7849\\
        Random &-& 0.3030 & 7.7388\\\hline
        Co-NavGPT \cite{yu2023Co-NavGPT} &\makecell{GPT-4\\GPT-4o} &\makecell{0.3209\\0.2896} &\makecell{6.1336\\7.6152}\\
    \bottomrule
    \end{tabular}
    \caption{Scene Graph Collaborative Exploration}
    \vspace{-1em}
    \label{tab:scene_graph}
\end{table}


\begin{table}[]
    \centering
    \small
    \begin{tabular}{c|cccc}
    \toprule
        Type &  SR & SPL & MPL & LPL    \\
    \midrule
        \makecell{Hierarchical interaction\\(VLM Explore)} & 0.00& 0.00&3.25& 48.65 \\\hline
        \makecell{Hierarchical interaction\\(VLM Explore+Act Prim)} & 0.00& 0.00&0.00 & 50.00 \\\hline
        \makecell{Horizontal interaction\\(VLM Zero-Shot)} & 0.00& 0.00&6.82& 49.52 \\
    \bottomrule
    \end{tabular}
    \caption{Open World Social Mobile Manipulation. The VLM here is GPT-4o \cite{achiam2023gpt4o}.}
    \label{tab:npc}
    \vspace{-1em}
\end{table}

\subsection{Metrics}
\begin{itemize}
    \item \textbf{Object Loco-Navigation Metrics.} We use common metrics in navigation tasks, including \textbf{SR} (Success Rate), \textbf{SPL} (Success weighted by Path Length), which is weighted by the ratio of the actual path length to the ground truth path length. Additionally, \textbf{NE} (Navigation Error), the distance to the target at the end of the navigation, to measure the agent's performance in terms of navigation success, efficiency, and other aspects.
    \item \textbf{Loco-Manipulation Metrics.} Similar to Object Loco-Navigation, we additionally include an evaluation to determine whether the agent can manipulate the specified object. The metrics include \textbf{SR}, \textbf{SRL}, and \textbf{NE} based on the entire process of navigation and manipulation.
    \item  \textbf{Scene Graph Collaborative Exploration Metrics.} We set the maximum exploration steps for the robot in the scene to 200. The ratio of the number of object instances discovered by the robot under this condition to the actual number of object instances in the scene is defined as the Semantic Exploration Rate (\textbf{SER}). Additionally, the Minimum Root Mean Square Error (\textbf{MRMSE}) between the centers of objects located by the robot and the actual objects is used to evaluate the efficiency and accuracy of the robot's exploration. 
     \item  \textbf{Open World Social Mobile Manipulation Metrics.} In this section, we use \textbf{SR} and \textbf{SPL} from the Loco-Manipulation Metrics as our evaluation metrics. In addition, we also evaluated the robot's minimum action path (\textbf{MPL}) and the longest action path (\textbf{LPL}), to measure the large model's perception of the robot's actions.
\end{itemize}
\vspace{-0.5em}
\subsection{Evaluation}
\begin{itemize}
\item  \noindent \textbf{Object Loco-Navigation.} 
For Object Loco-Navigation, LLM-Based Ins following with GPT-4o \cite{achiam2023gpt4o} achieved excellent performance. As Table \ref{tab:obj-loco} shows, with the help of the navigation interface, SR reached 90.82\%, SPL reached 90.82\%, and NE reached 1.0. The failure cases were due to the agent failing to reach the position where the object was within a 60-degree horizontal view, with a wall or obstacle blocking the view. 
It is worth noting that between Qwen and Chat-GLM4, Qwen produced more stable actions, but its accuracy in generating actions was suboptimal, making it ineffective at precisely locating the specified object in the designated area. On the other hand, while Chat-GLM4's stability was lower than Qwen's, its action accuracy was relatively higher. 
For VLMs, the performance of all VLM models is similarly low, demonstrating that under zero-shot settings, VLMs still struggle to achieve the goal solely through direct observation and action generation.

\item  \noindent \textbf{Loco-Manipulation.} For navigation manipulation tasks, the differences between models were even more pronounced. These tasks require precise judgment of manipulation actions and involve multi-stage processes, emphasizing the importance of action accuracy. As shown in Table \ref{tab:loco-maniplation}, among LLMs, GPT-4o maintained the highest performance. However, due to its higher action accuracy, Chat-GLM4 achieved a significantly better success rate compared to Qwen. Mobile manipulation is equally challenging for VLMs. VLMs not only struggle to reach the target but also find it difficult to determine the boundaries of whether an object can be grasped. This poses significant challenges for VLMs.

\item  \noindent \textbf{Scene Graph Collaborative Exploration.} We conducted additional experiments on Co-NavGPT using GPT-4, as the original experiments were based on the more commonly used GPT-4-turbo. As shown in Table \ref{tab:scene_graph}, the results showed that GPT-4 performed the best, possibly due to the design of the prompts used.

\item  \noindent \textbf{Open World Social Mobile Manipulation.} We noticed that using VLM to directly output discrete actions in Hierarchical Interaction resulted in a success rate of 0 for the robot, we have now incorporated additional action primitives. For example, actions like $<\textit{walk}>$, which allows the robot to move to a specific object location on the known map, and $<\textit{pick}>$, which enables the robot to directly grab the target object from its current viewpoint using planning. We then conducted further planning experiments using VLM. However, from Table \ref{tab:npc} the final results still yielded a success rate of 0. Analyzing the constructed maps, we found that since we used the results from Benchmark 3, most of the maps were built using semantic information, which was often too coarse. As a result, the object instances corresponding to the tasks might not have appeared in the constructed maps, or the parsed positions had large discrepancies from the actual locations.

\end{itemize}

\section{Conclusions}
\label{sec:conclusions}
In this paper, we present InfiniteWorld, a unified and scalable simulation framework for vision-language robotic interaction, which includes unlimited interactable physics assets and a comprehensive free-form robotic interaction benchmark.
We aim to provide the community with a comprehensive simulation platform that includes a variety of rich 3D asset construction interfaces and supports unlimited expansion of scenarios to alleviate the plight of the lack of high-quality embodied assets. At the same time, we build a benchmark for robot social interaction in open scenarios to comprehensively evaluate the capabilities of embodied agents in terms of perception, planning, execution, and interaction.
\clearpage
{
    \small
    \bibliographystyle{ieeenat_fullname}
    \bibliography{main}
}

\clearpage
\setcounter{page}{1}
\maketitlesupplementary

In the supplementary material, we present more details about the simulator in Section \ref{app:SimulationDetails}. In Section \ref{app:experimentalDetails}, we present more experimental details and results.

\section{Simulation Details}
\label{app:SimulationDetails}

\subsection{Depth-Prior-Constrained Real2Sim Pipeline}
\label{app:Real2SimPipeline}
Specifically, Our 3D scene reconstruction pipeline includes the entire process from photographic data to accurate and visually coherent models. Its main steps are as follows:
\begin{itemize} \item \textbf{SfM.} The process begins with colmap-glomap \cite{pan2024glomap}, an SfM approach that estimates camera parameters and produces a sparse point cloud.



\item \textbf{Novel View Synthesis (NVS) \& Meshing.} NVS is achieved through the improved PGSR \cite{chen2024pgsr}, after which mesh extraction is conducted with Truncated Signed Distance Function (TSDF) \cite{newcombe2011kinectfusion} and the Marching Cubes algorithm \cite{lorensen1998marching}.

\item \textbf{Z-Axis Alignment.} To ensure correct vertical orientation, we employ the Random Sample Consensus (RANSAC) \cite{fischler1981random} algorithm to detect and align the dominant plane and rotate the whole scene for z-axis alignment.

\item \textbf{Denoising.} Using a connectivity-cluster approach, we effectively filter noise, setting a threshold to remove extraneous points from high spatial areas. This step reduces model complexity and enhances visual clarity.

\item \textbf{Hole-Filling.} Small gaps in the mesh are closed with PyMeshFix \cite{attene2010lightweight}, which preserves the structural continuity of the model and maintains its overall integrity.

\item \textbf{Recoloring.} To restore color lost during hole-filling, we map colors from the original images to the mesh vertices using KDTree \cite{friedman1977algorithm}, ensuring consistent color information across the model.

\item \textbf{Simplification.} 
Finally, PyMeshLab \cite{PyMeshLab} is used to reduce vertex density to optimize the model size, which minimizes complexity while retaining essential geometry.
\end{itemize}

\begin{figure}[htb]
    \centering
    \addtolength{\tabcolsep}{-6.5pt}
    \footnotesize{
        \setlength{\tabcolsep}{1pt} 
        \begin{tabular}{p{8.2pt}cc}
            & \textbf{w.o. post-processing} & \textbf{w. post-processing} \\
            
            \raisebox{23pt}{\rotatebox[origin=c]{90}{Cupboard}}&
            \includegraphics[width=0.45\linewidth]{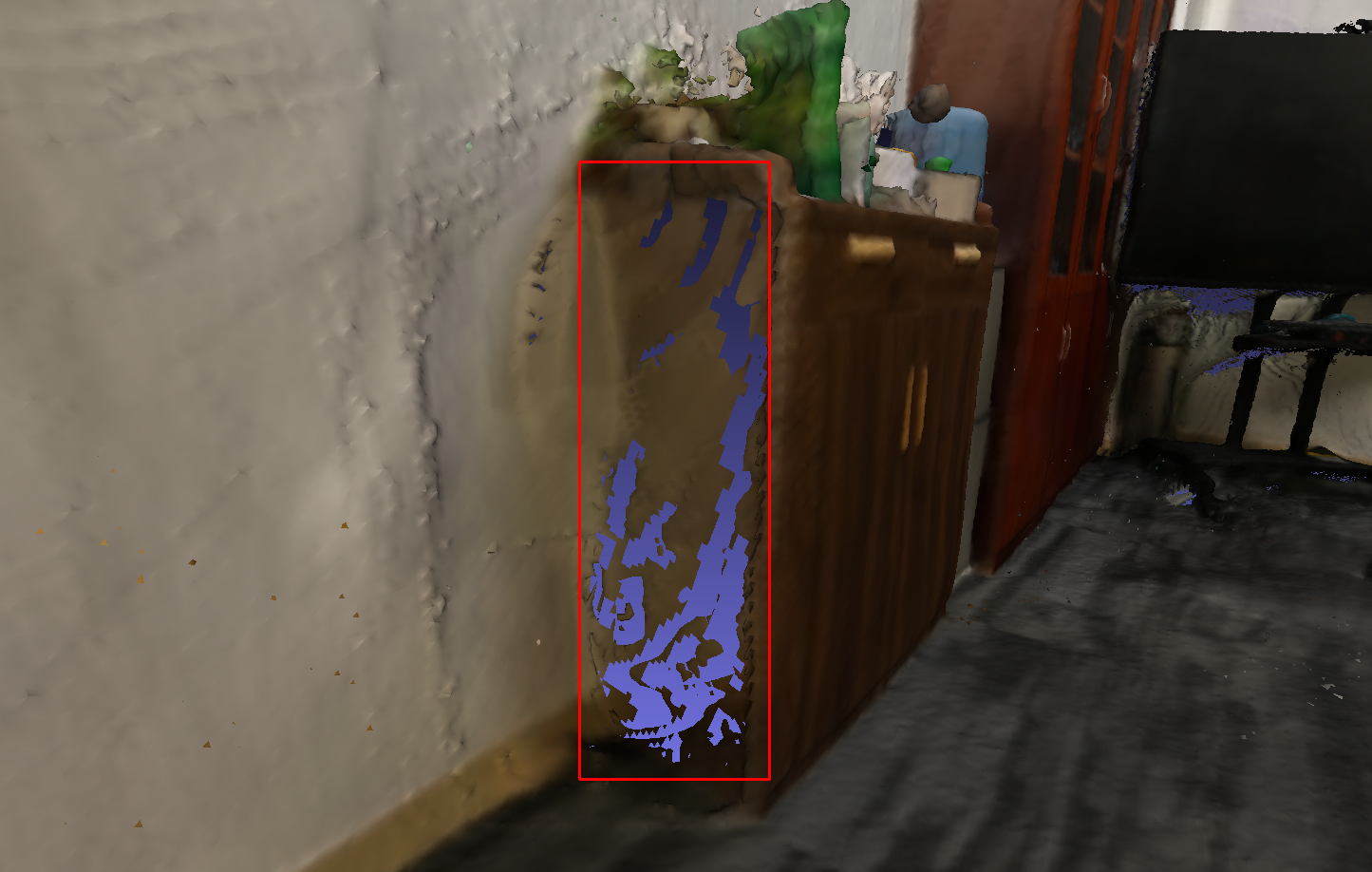} &
            \includegraphics[width=0.45\linewidth]{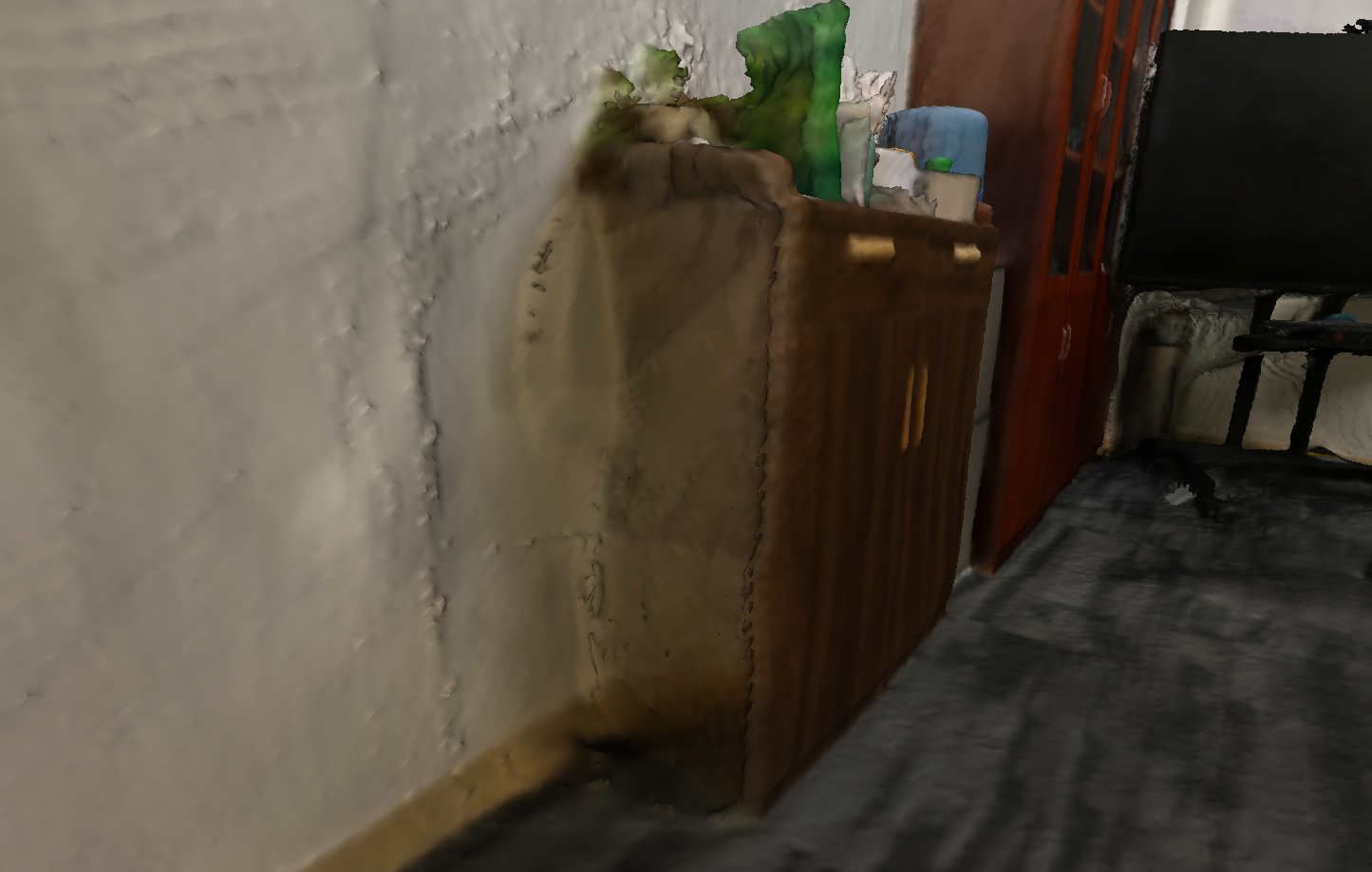}
            \\

            \raisebox{23pt}{\rotatebox[origin=c]{90}{Sofa}}&
            \includegraphics[width=0.45\linewidth]{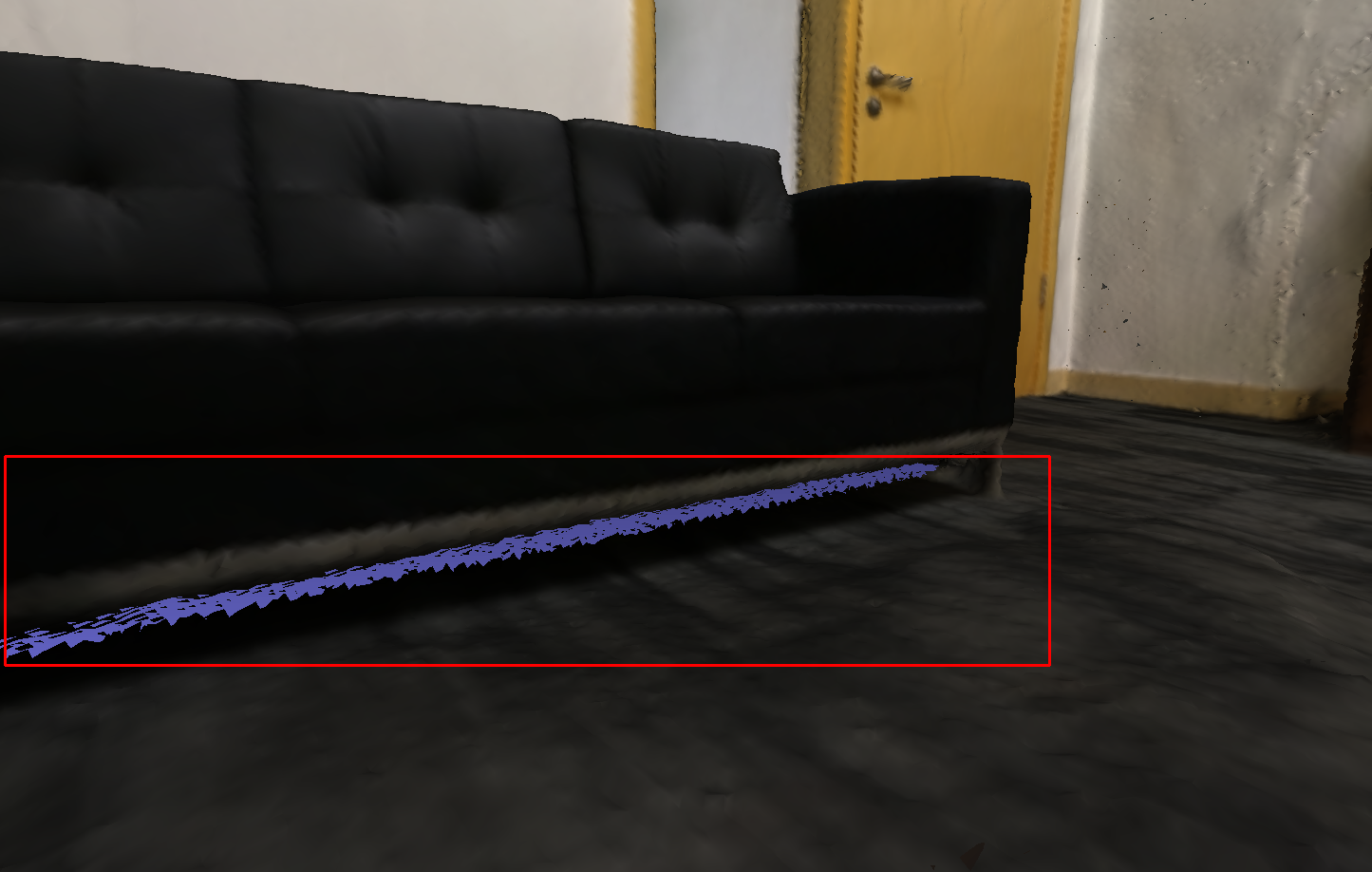} &
            \includegraphics[width=0.45\linewidth]{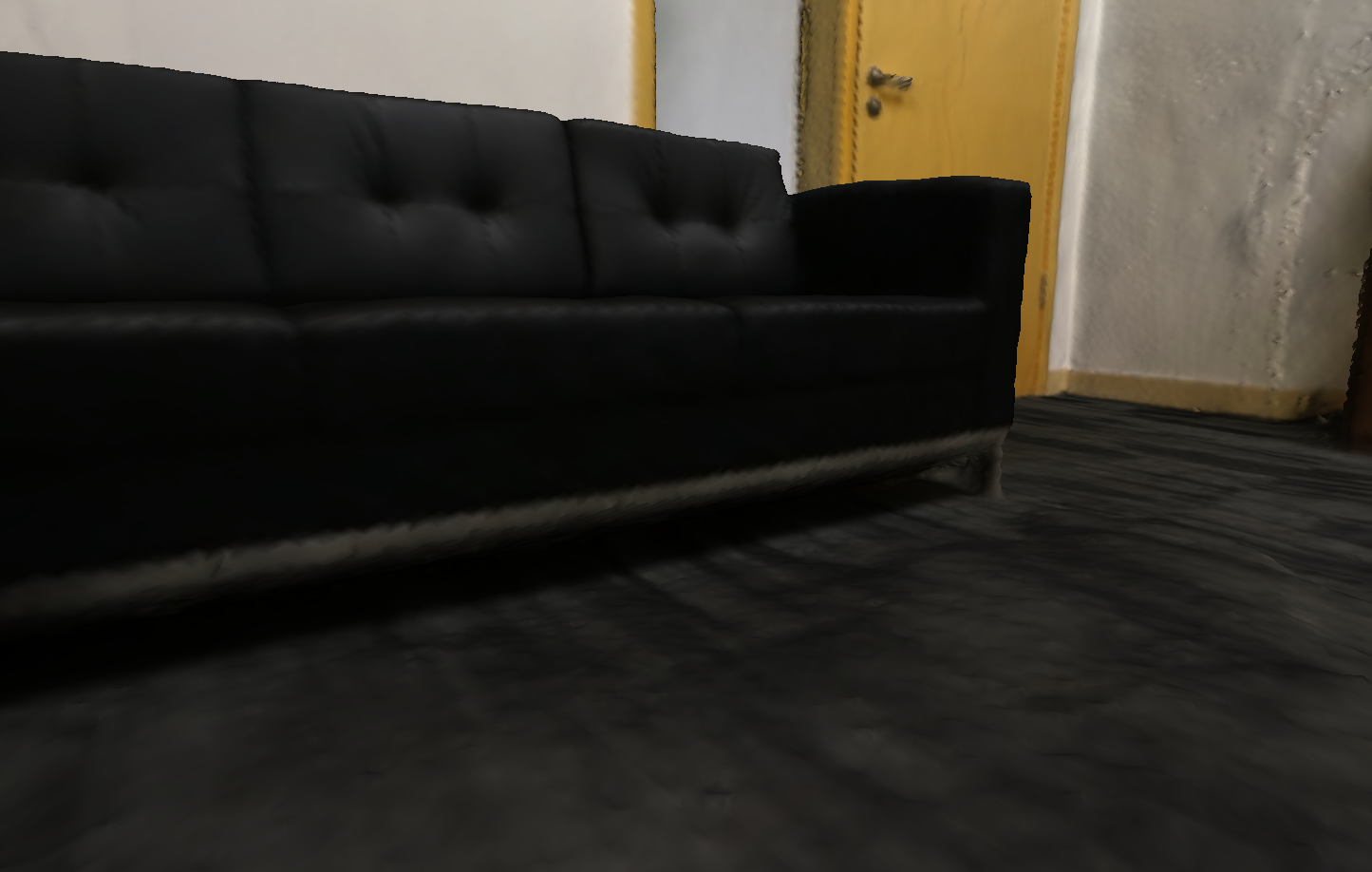}
            \\
            
            \raisebox{23pt}{\rotatebox[origin=c]{90}{Table}}&
            \includegraphics[width=0.45\linewidth]{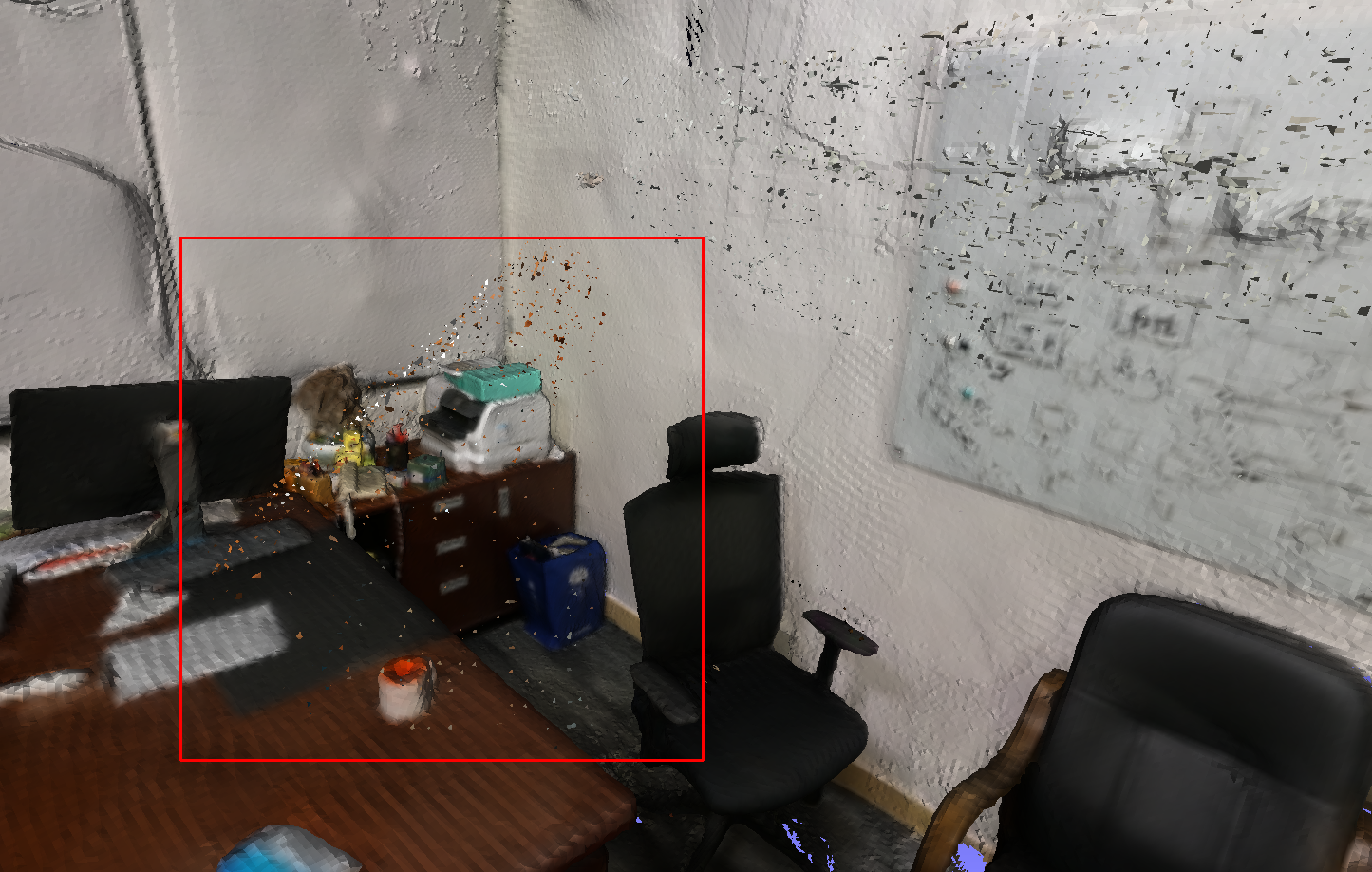} &
            \includegraphics[width=0.45\linewidth]{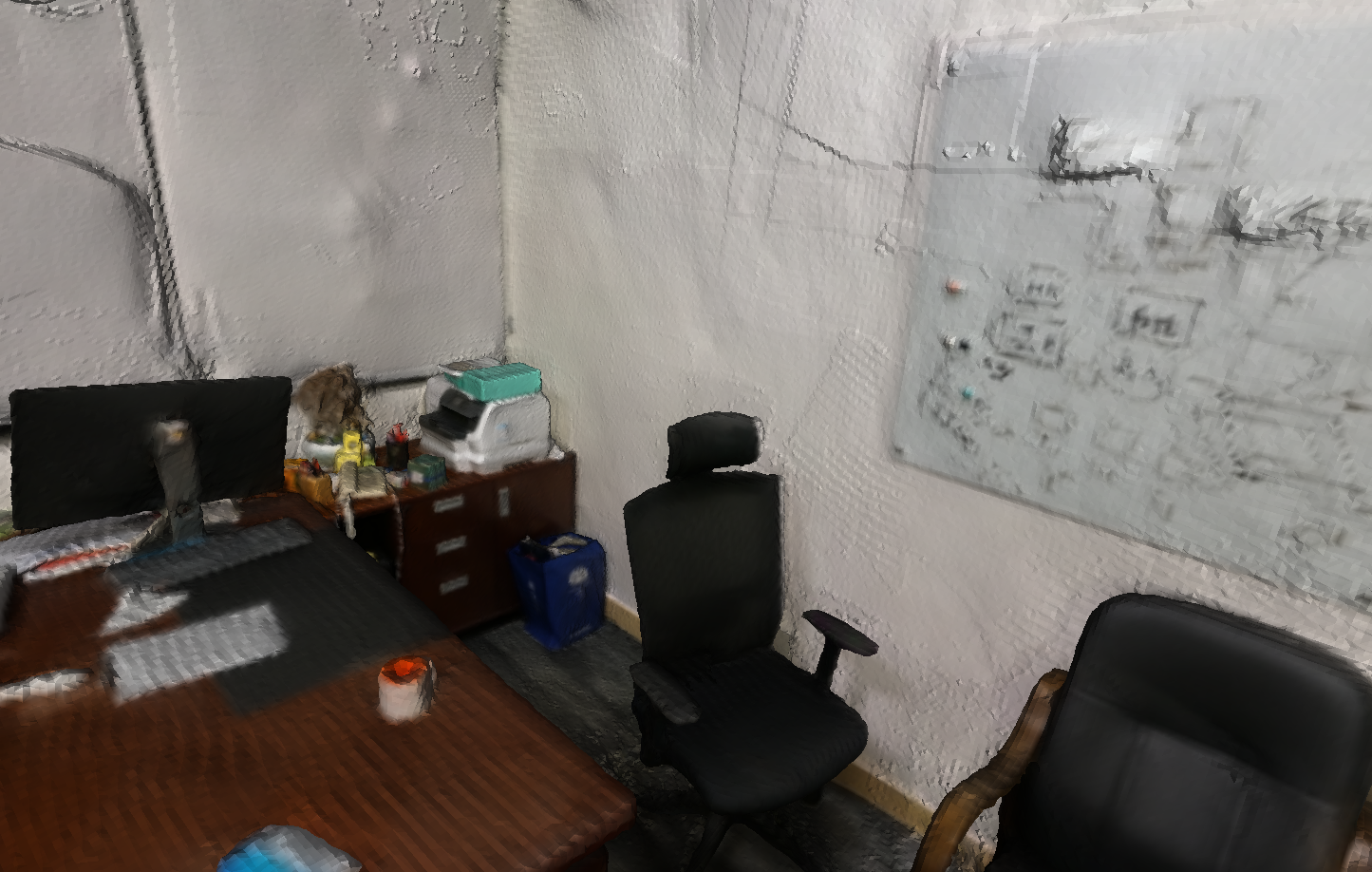} 

        \end{tabular}
    }
    \vspace{-5pt}
	\caption{Comparative of reconstruction results with and without post-processing. Visualizing the results shows that our post-processing method is very effective in resolving holes and removing floating meshes in the scene, such as the gaps around cabinets, sofas, and the floaters above the table in the red bounding boxes.}
    \label{fig:compare_of_postp}
\end{figure}

In addition, post-processing workflows for 3D scene reconstruction are critical for refining 3D models and enhancing their usability in simulated environments.
This process encompasses key refinements that address axis alignment, noise reduction, surface continuity, and model size. 
As shown in Fig.~\ref{fig:compare_of_postp}, the real-world appearance of the model undergoes significant improvement after post-processing, as compared to the initial results.

\begin{table*}[t]
\newcommand{\greencell}[1]{\cellcolor{green!20}{#1}}
\newcommand{\redcell}[1]{\cellcolor{red!20}{#1}}
\begin{tabular}{l|ccccccc}
\hline
Tool & 3D BAT & SAnE & SUSTech & Label & ReBound & Xtreme1 & Annot8-3D \\
 & \cite{zimmer20193d} & \cite{9143095} & POINT\cite{li2020sustech} & Cloud\cite{sager2021labelcloud} & \cite{rebound} & \cite{Xtreme1} & \textbf{(Ours)} \\
\hline
Year & 2019 & 2020 & 2020 & 2021 & 2023 & 2023 & 2024 \\
2D/3D cam.+LiDAR fusion & \greencell{\checkmark} & \redcell{-} & \greencell{\checkmark} & \greencell{\checkmark} & \greencell{\checkmark} & \greencell{\checkmark} & \greencell{\checkmark} \\
AI-assisted labeling & \greencell{\checkmark} & \greencell{\checkmark} & \greencell{\checkmark} & \redcell{-} & \greencell{\checkmark} & \greencell{\checkmark} & \greencell{\checkmark} \\
Label custom attributes & \redcell{-} & \redcell{-} & \greencell{\checkmark} & \redcell{-} & \greencell{\checkmark} & \greencell{\checkmark} & \greencell{\checkmark} \\
HD Maps & \redcell{-} & \redcell{-} & \greencell{\checkmark} & \redcell{-} & \redcell{-} & \greencell{\checkmark} & \greencell{\checkmark} \\
Web-based & \greencell{\checkmark} & \redcell{-} & \greencell{\checkmark} & \redcell{-} & \redcell{-} & \greencell{\checkmark} & \greencell{\checkmark} \\
3D navigation & \greencell{\checkmark} & \greencell{\checkmark} & \greencell{\checkmark} & \redcell{-} & \greencell{\checkmark} & \greencell{\checkmark} & \greencell{\checkmark} \\
3D transform controls & \greencell{\checkmark} & \greencell{\checkmark} & \greencell{\checkmark} & \redcell{-} & \greencell{\checkmark} & \greencell{\checkmark} & \greencell{\checkmark} \\
Side views (top/front/side) & \greencell{\checkmark} & \greencell{\checkmark} & \greencell{\checkmark} & \redcell{-} & \greencell{\checkmark} & \greencell{\checkmark} & \greencell{\checkmark} \\
Perspective view editing & \greencell{\checkmark} & \greencell{\checkmark} & \greencell{\checkmark} & \greencell{\checkmark} & \greencell{\checkmark} & \greencell{\checkmark} & \greencell{\checkmark} \\
Orthographic view editing & \greencell{\checkmark} & \greencell{\checkmark} & \redcell{-} & \redcell{-} & \greencell{\checkmark} & \greencell{\checkmark} & \greencell{\checkmark} \\
Object coloring & \greencell{\checkmark} & \greencell{\checkmark} & \greencell{\checkmark} & \redcell{-} & \greencell{\checkmark} & \greencell{\checkmark} & \greencell{\checkmark} \\
Offline annotation support & \redcell{-} & \redcell{-} & \redcell{-} & \redcell{-} & \redcell{-} & \redcell{-} & \greencell{\checkmark} \\
Multi-stage Annotation & \redcell{-} & \redcell{-} & \redcell{-} & \redcell{-} & \redcell{-} & \redcell{-} & \greencell{\checkmark} \\
Physical Attributes Labeling & \redcell{-} & \redcell{-} & \redcell{-} & \redcell{-} & \redcell{-} & \redcell{-} & \greencell{\checkmark} \\
\hline
\end{tabular}
\caption{Comparison of 3D annotation tools. \textcolor{green}{$\bigcirc$} Feature provided \textcolor{red}{$\bigcirc$} Feature not provided}
\label{tab:Annot8-3D}
\end{table*}

\subsection{Details of Annot8-3D}
\label{appendix:annot8-3d}
Annot8-3D supports common 3D point cloud formats, with a comprehensive attribute schema capturing physical and semantic properties essential for robotics applications. These attributes span multiple categories: essential properties including unique identifiers and collision characteristics; manipulation-related features such as friction coefficients, manipulability flags, and instance segmentation; navigation-centric data including position coordinates, room assignments, and orientation relative to traversable space; and optional descriptors covering semantic labels and appearance characteristics.

Table \ref{tab:Annot8-3D} presents a detailed feature comparison between Annot8-3D and six 3D annotation tools from 2019 to 2024. The comparison reveals three distinct categories of features: (1) Common features widely supported across tools, including perspective view editing, which is universally available, and 3D navigation and transformation controls, supported by most platforms; (2) Partially supported features, such as 2D/3D camera and LiDAR fusion, AI-assisted labeling, and custom attribute labeling, which are present in some but not all tools; and (3) Unique features exclusive to Annot8-3D, specifically offline annotation support, multi-stage annotation pipeline, and physical attributes labeling. While newer tools like ReBound and Xtreme1 (both from 2023) have incorporated advanced features such as AI-assisted labeling and custom attributes, Annot8-3D further extends these capabilities through its comprehensive physical attribute schema and multi-stage annotation approach. Additionally, it maintains compatibility with essential features present in earlier tools while introducing novel functionalities for robotics applications.

\begin{figure*}
    \centering
    \includegraphics[width=1\linewidth]{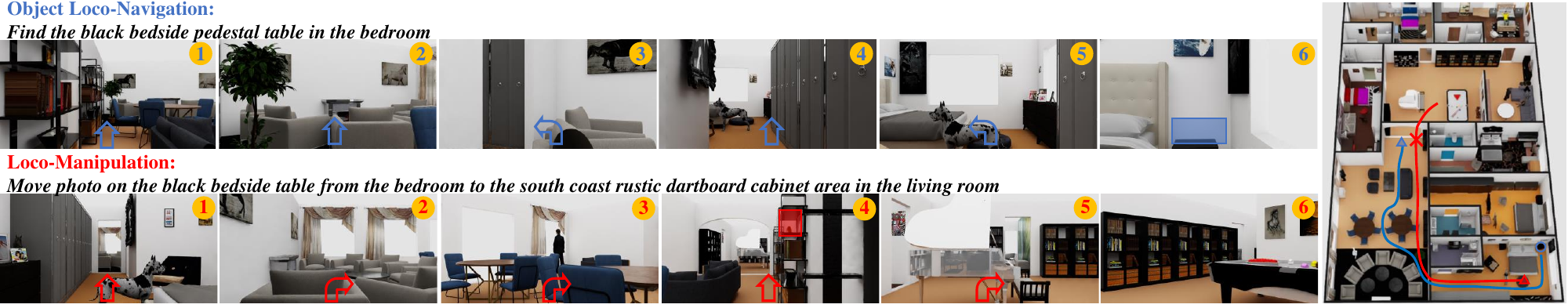}
    \vspace{-1em}
    \caption{Visualization of Benchmark 1 and Benchmark 2.}
    \label{fig:traj}
\end{figure*}

\subsection{Unified 3D Asset}
\label{app:asset}
Overall, Table \ref{tab:obj_assets} summarizes the statistics of these object asset datasets. 
The corresponding scene asset statistics are shown in Table \ref{tab:scene_assets}.

\begin{table*}[!ht]
    
    \centering
    \small
    \begin{tabular}{l|l l l l c c } 
        \toprule  
         Dataset & Num. &Type & Classes &Format & Texture & Interactive  \\
         \midrule
         3D-Front \cite{3DFront}& 5,172 & Indoor furniture& 21 &\textit{.obj} & \Checkmark & \XSolidBrush  \\
         Objaverse \cite{deitke2024objaverse}&  4,042,937 & Small objects& 940 &\textit{.pkl}&  \Checkmark  & \XSolidBrush  \\
         ClothesNet \cite{zhou2023clothesnet}&  3,051 &Soft clothing objects& 11 &\textit{.obj}+\textit{.urdf}  & \Checkmark & \Checkmark  \\
         PartNet-mobility \cite{mo2019partnet}&  26,671 & Articulated rigid objects& 24 &\textit{.obj}+\textit{.urdf}  & \Checkmark & \Checkmark  \\
        \bottomrule
    \end{tabular}
    \caption{Statistical information of the object asset.
    }
    \label{tab:obj_assets}
\end{table*}

\begin{table}[t]
    
    \centering
    \small
    \begin{tabular}{l|l l c c } 
        \toprule  
         Dataset & Num. &Format  & Texture & Interactive  \\
         \midrule
         HM3D \cite{HM3D}& 1,000  & \textit{.glb} & \Checkmark & \XSolidBrush  \\
         
         HSSD \cite{10657917hssd}&  120  &\textit{.glb}&  \Checkmark  & \Checkmark  \\
         
         Replica \cite{straub2019replica}&  18 &\textit{.ply}  & \Checkmark &  \XSolidBrush \\
        Scannet \cite{dai2017scannet} &  1513 &\textit{.ply}  & \Checkmark &  \XSolidBrush \\
         
      
        \bottomrule
    \end{tabular}
    \caption{Statistical information of the scene asset.}
    \label{tab:scene_assets}
\end{table}

\section{Experimental Details}
\label{app:experimentalDetails}
\subsection{Task Setting}
\label{app:TaskSettingDetails}
We provide simulation assistance to help users complete various customized tasks on InfiniteWorld Simulation.
\begin{itemize}
\item  \noindent \textbf{Occupy Map.} For each scene, we generate an occupy map, a two-dimensional grid map used for embodied agent navigation. The occupy map projects the scene along the \textit{z}-axis onto the \textit{xy}-plane and divides the scene into three areas: “free”, “obstacle”, and “unknown”. The agent can move in the “free” area and will be blocked by “obstacles”. Based on the occupy map, the agent can plan its movement within the scene.

\item  \noindent \textbf{Path Follower.} We provide a path follower for agents that enables point-to-point path planning. We utilize the D* Lite algorithm based on the occupancy map to optimally find paths while avoiding obstacles. In object loco-navigation tasks, the coordinates of objects often lie within “obstacle” areas. When the target point is in these illegal zones, the path follower will identify the nearest non-colliding point on the occupancy map as an alternative target point, ensuring the feasibility of the navigation path. Embodied agents can directly use the path follower to achieve object loco-navigation with the help of scene semantics, or just use the path follower as a supervisory signal for imitation learning.

\item  \noindent\textbf{Physical Manipulation.} We provide joint-based robot arm control for embodied agents. The agents can achieve forward control by directly providing target joint angles or achieve inverse control by specifying the end effector's pose through inverse kinematics solving. The end effector of the robot arm will interact with objects based on physics and return real-time physical feedback.

\item  \noindent\textbf{Adhesion.} We provide an adhesion interface for embodied agents. Unlike physical manipulation, the adhesion interface does not require the end effector to physically interact with the object. When the object is within a certain range of the end effector, the adhesion interface can directly attach the object to the end effector, allowing it to move with the agent until the adhesion is released. This eliminates the need for the agent to consider grasping poses and trajectories in physical manipulation.

\end{itemize}

\subsection{Benchmark Setting}
\label{app:benchmarkSetting}
\textbf{Benchmark 1: Object Loco-Navigation.} The basic format of the task is “Find an $<object>$ in $<room>$." \\
\textbf{Benchmark 2: Loco-Manipulation.} The basic format of the task is “take the $<object\ 1>$ in $<room\ 1>$ to $<object\ 2>$ in $<room\ 2>$." The agent needs to navigate to the vicinity of $<object 1>$ and accurately locate and grasp the object using a robotic arm. After moving $<object\ 1>$ close to $<object\ 2>$, the agent needs to maneuver the robotic arm to place $<object\ 1>$ in the specified position.\\
\textbf{Benchmark 3: Scene Graph Collaborative Exploration. } The most basic requirement of the task is “Please explore the entire scene as quickly as possible”, and the robot uses an algorithm to record the spatial position of each scene instance object.
\\
\textbf{Benchmark 4: Open World Social Mobile Manipulation.} Similar to Benchmark2, the task format assigned to the nth robot is “$<robot\ n>$, please take the $<object\ n1>$ in $<room\ n1>$ to $<object\ n2>$ in $<room\ n2>$.” In Hierarchical Interaction, the robot utilizes information from previously constructed maps to accomplish tasks. Specifically, the map information is formatted as prompts and input into a large model for planning. In contrast, in Horizontal Interaction, robots operate independently without direct information sharing. Communication is only enabled when the distance between robots reaches a certain threshold, allowing information exchange through inquiry actions, such as obtaining map information constructed by another robot.

\end{document}